\documentclass[11pt]{article}

\usepackage[final]{acl}

\usepackage{times}
\usepackage{latexsym}
\usepackage[T1]{fontenc}
\usepackage[utf8]{inputenc}
\usepackage{microtype}
\usepackage{inconsolata}
\usepackage{graphicx}
\usepackage{booktabs}
\usepackage{amsmath}
\usepackage{amssymb}
\usepackage{amsthm}
\usepackage{xcolor}
\usepackage{enumitem}
\usepackage{float}
\usepackage{longtable}
\usepackage{array}
\usepackage{tabularx}
\usepackage{threeparttable}
\usepackage{siunitx}
\usepackage{listings}
\usepackage{pifont}
\usepackage{tikz}
\usetikzlibrary{arrows.meta,positioning,shapes.geometric,fit,backgrounds,calc,decorations.pathreplacing}
\usepackage[most]{tcolorbox}
\definecolor{cbblue}{HTML}{0072B2}
\definecolor{cborange}{HTML}{E69F00}
\definecolor{cbgreen}{HTML}{009E73}

\newcommand{\benchmarkname}{EComAgentBench}

\newcommand{\TotalSamples}{921}
\newcommand{\ValidSamples}{662}
\newcommand{\GenerationPassRate}{71.9}
\newcommand{\IntentCount}{8}
\newcommand{\ToolCount}{10}
\newcommand{\RubricCount}{6645}
\newcommand{\RubricTypeCount}{6}
\newcommand{\AvgRubrics}{10.04}
\newcommand{\QueryRubricCount}{3993}
\newcommand{\PersonaRubricCount}{1326}

\newcommand{\QueryRubricRatio}{60.1}
\newcommand{\PersonaRubricRatio}{19.9}
\newcommand{\ClarificationRubricRatio}{19.9}
\newcommand{\BestAccuracy}{57.1}
\newcommand{\WorstAccuracy}{19.5}
\newcommand{\JudgeModel}{gemini-3.1-pro-preview}
\newcommand{\CaseExampleModel}{gpt-5.4}

\graphicspath{{figures/}}

\newcolumntype{Y}{>{\raggedright\arraybackslash}X}
\newcommand{\yes}{\textcolor{black!80}{\ding{51}}}
\newcommand{\no}{\textcolor{black!55}{\ding{55}}}
\newcommand{\partialmark}{\textcolor{black!70}{$\sim$}}

\lstdefinelanguage{json}{
  basicstyle=\ttfamily\footnotesize,
  showstringspaces=false,
  breaklines=true,
  morestring=[b]",
  stringstyle=\color{green!40!black},
  literate=
   *{0}{{{\color{blue}0}}}{1}
    {1}{{{\color{blue}1}}}{1}
    {2}{{{\color{blue}2}}}{1}
    {3}{{{\color{blue}3}}}{1}
    {4}{{{\color{blue}4}}}{1}
    {5}{{{\color{blue}5}}}{1}
    {6}{{{\color{blue}6}}}{1}
    {7}{{{\color{blue}7}}}{1}
    {8}{{{\color{blue}8}}}{1}
    {9}{{{\color{blue}9}}}{1}
    {:}{{{\color{red!70!black}:}}}{1}
    {,}{{{\color{red!70!black},}}}{1}
    {\{}{{{\color{red!70!black}\{}}}{1}
    {\}}{{{\color{red!70!black}\}}}}{1}
    {[}{{{\color{red!70!black}[}}}{1}
    {]}{{{\color{red!70!black}]}}}{1},
}

\title{\benchmarkname{}: Benchmarking Shopping Agents on Long-Horizon Tasks with Distributed Hidden Intent%
\thanks{An earlier version of this paper appeared as arXiv:2606.17698.}}

\author{Zeyao Du, Tong Li, Haibo Zhang\thanks{Corresponding author.} \\
  Shopee \\
  \texttt{\{zeyao.du,litonglt,peter.wu\}@shopee.com}}

\begin{document}
\maketitle

\begin{abstract}
As LLM-based shopping agents enter production, existing benchmarks fail to capture how a shopper's requirements arrive: stated implicitly in the query, recorded in a profile, or revealed only when the right question is asked. Benchmarks that expose full intent upfront and grade only the final choice can neither pose this long-horizon challenge nor explain which requirement an agent missed. To address this gap, we introduce \benchmarkname{}, a benchmark of \ValidSamples{} tasks grounded in real Amazon products and reviews. Each task scatters these requirements across a visible query, a tool-gated profile, and scripted clarification; an agent must uncover hidden intent, verify candidates against attributes and review evidence, and commit to a single product within 100 tool calls. Moreover, typed, source-tagged rubrics grade every task, attributing each failure to a requirement and its source. Construction is automated yet reliable, with every answer fixed in code before any text is generated and every sample validated. Our evaluation of seven models reveals that even the strongest attains only \BestAccuracy{}\% overall accuracy, and rubric satisfaction degrades from visible to hidden sources. Overall, we believe \benchmarkname{} will serve as a reproducible foundation for moving shopping agents from single-query search toward dependable assistance over long horizons.

\end{abstract}

\section{Introduction}

E-commerce shopping agents are moving from single-query product lookup to assistants that serve a shopper over many tool-using steps. Selecting a model for deployment therefore requires a benchmark that separates models, explains where each fails, and reflects how shopping requests actually arrive, while being constructed and graded reliably enough to trust the resulting ranking. Prior shopping and web-agent benchmarks established that realistic evaluation is a \emph{long-horizon} interaction problem over tools and partial information, not single-shot question answering~\citep{webshop2022,mind2web2023,webarena2024,assistantbench2024}.

We identify three weaknesses in how shopping agents are evaluated today, each motivating one of our contributions. \textbf{(i)~Coarse, non-diagnostic scoring:} prior shopping benchmarks score an answer only by exact product match or by feature overlap with the target~\citep{webshop2022,shoppingbench2026}. No such score can say which requirement of a many-step task was missed, nor where it came from: the query, the profile, or a follow-up question. \textbf{(ii)~Little hidden intent:} real shoppers state some needs implicitly, keep others in their profile, and mention some only when asked. Most benchmarks, however, place all requirements in the opening query, leaving the long trajectory nothing new to discover; the few with personalization expose the profile directly or simulate the user. \textbf{(iii)~Unreliable construction:} long-horizon tasks are costly to author and noisy to grade. Existing benchmarks each forgo at least one of validation, automation, or real products; many further simulate the user or leave the judge unconstrained, adding variance orthogonal to the capability under test.

To address all three, we present \benchmarkname{}, a single-product shopping benchmark on a real Amazon Reviews 2023 catalog~\citep{amazonreviews2024} that splits each shopper's requirements across three sources: a \emph{visible query} that states some needs only implicitly, a \emph{hidden persona} read only via the \texttt{get\_user\_profile} tool, and \emph{hidden clarification} obtained only by asking a question the \texttt{ask\_user} tool recognizes. Because two sources sit behind tools, the full intent cannot be read in any single step; it must be assembled across the episode (Figure~\ref{fig:overview}; worked example in Section~\ref{sec:task_env}). We grade each episode with checks we call \emph{rubrics}, each labeled by type and by source (query, persona, or clarification), so a failed check identifies which requirement was missed and where. Construction is automated yet reliable: each answer is fixed in code before any text is generated, clarification is scripted rather than simulated, and every sample undergoes multi-stage validation.

We organize the paper around three questions and answer each with evidence from seven representative models evaluated under a shared environment and judge:
\begin{itemize}[leftmargin=1.2em,itemsep=2pt,topsep=2pt]
  \item \textbf{RQ1: Does the benchmark separate models and localize failures in long-horizon tasks by requirement and source?} Overall accuracy spans \WorstAccuracy{}--\BestAccuracy{}\%, and the typed, source-tagged rubrics pinpoint which requirements each model misses and where they originate.
  \item \textbf{RQ2: Does hidden, distributed intent construct a realistic long-horizon task?} Splitting requirements across the query, the profile, and clarification mirrors how shoppers express intent, and is consistently harder for every model, isolating a capability that query-visible evaluation misses.
  \item \textbf{RQ3: Can long-horizon evaluation be trusted and scaled?} Construction is automated yet validated over a real catalog, and the judge errs rarely: it rejects a correct answer at most once per model, and accepts no incorrect answer in an independent audit; the category-agnostic pipeline scales by sampling more products, not re-annotating.
\end{itemize}

In short, \benchmarkname{} shows how to construct long-horizon shopping tasks (hidden, distributed intent), diagnose their failures (typed, source-tagged rubrics), and keep both reproducible (automated, validated construction): a combination no prior shopping benchmark provides.

\section{Related Work}

\paragraph{E-commerce shopping benchmarks.} WebShop~\citep{webshop2022} grounds agents in real shopping pages but focuses on browser interaction, not on recovering hidden requirements. Recent shopping benchmarks differ in both \emph{what intent they hide} and \emph{how they are built}. ShoppingBench~\citep{shoppingbench2026} keeps intent essentially query-visible and auto-generates tasks without independent validation; ShopSimulator~\citep{shopsimulator2026} adds personalization through a simulated user but relies on costly manual annotation; DeepPlanning~\citep{deepplanning2026} offers verifiable-constraint long-horizon planning, but on synthetic products rather than a real catalog. However, none reveals where a failed requirement came from, even when granting partial credit. Together these gaps motivate \benchmarkname{}: a task whose intent is split across three sources, two gated behind tools, so that recovering it spans a long horizon; typed, source-tagged rubrics that attribute each failure to a requirement and its source; and a pipeline that is at once automated, validated, and built on a real Amazon catalog (Table~\ref{tab:comparison}).

\begin{table*}[t]
\centering
\begin{threeparttable}
\renewcommand{\arraystretch}{1.0}
\setlength{\tabcolsep}{6pt}
\footnotesize
\begin{tabular}{l cccc}
\toprule
\textbf{Capability} & \textbf{ShoppingBench} & \textbf{ShopSimulator} & \textbf{DeepPlanning} & \textbf{Ours} \\
\midrule
\multicolumn{5}{@{}l}{\textit{Explains failures\, (RQ1)}} \\
\quad Per-requirement (rubric) scoring\tnote{c} & \partialmark & \partialmark & \partialmark & \yes \\
\quad Shows which source failed                & \no  & \no          & \no  & \yes \\
\midrule
\multicolumn{5}{@{}l}{\textit{Realistic hidden intent\, (RQ2)}} \\
\quad Tool-gated (hidden) intent             & \no  & \partialmark & \no  & \yes \\
\quad Agent must ask for missing info        & \no  & \yes         & \no  & \yes \\
\midrule
\multicolumn{5}{@{}l}{\textit{Reliable, automated build\, (RQ3)}} \\
\quad Real product catalog                  & \yes & \yes         & \no          & \yes \\
\quad Automated construction                & \yes & \no          & \yes         & \yes \\
\quad Validated after building              & \no  & \yes         & \yes         & \yes \\
\bottomrule
\end{tabular}
\begin{tablenotes}[flushleft]\footnotesize
\item \yes~supported; \partialmark~partial; \no~not supported.
\item[c] The score breaks down to individual requirements; \partialmark{}~=~breaks down only to product-feature parts or task checkpoints.
\item Benchmarks: ShoppingBench~\citep{shoppingbench2026}, ShopSimulator~\citep{shopsimulator2026}, DeepPlanning~\citep{deepplanning2026}; VitaBench~\citep{vitabench2025} is discussed as evaluator-design inspiration rather than a primary comparison target.
\end{tablenotes}
\caption{\benchmarkname{} versus recent e-commerce shopping benchmarks, grouped by its three goals: explaining failures (RQ1), realistic hidden intent (RQ2), and a reliable, automated build (RQ3). \benchmarkname{} is the only one that meets all three; each prior benchmark already misses at least one capability in the reliability group (ShoppingBench has no task validation, ShopSimulator is hand-annotated, DeepPlanning uses synthetic products).}
\label{tab:comparison}
\end{threeparttable}
\end{table*}

\paragraph{Web-agent and interactive evaluation.} Mind2Web~\citep{mind2web2023}, WebArena~\citep{webarena2024}, VisualWebArena~\citep{visualwebarena2024}, and AssistantBench~\citep{assistantbench2024} showed that realistic agent evaluation should capture how an agent gathers information and uses its environment, not just the final answer; \benchmarkname{} focuses this idea on shopping, where intent must be recovered from the query, the profile, and clarification.

\paragraph{Evaluator design and rubric decomposition.} VitaBench~\citep{vitabench2025} pioneered rubric-based decomposition of complex agent trajectories; we extend this approach with rubrics that are \emph{typed and source-tagged}. Work on LLM-as-judge supports model-based grading but cautions against unconstrained judge scoring~\citep{zheng2023judging,wang2024fair,chiang2023alternative}, motivating a design in which the judge's targets are fixed in code and tied to typed rubrics rather than left to free-form judgment.

\section{Task and Environment}
\label{sec:task_env}

We now formalize this task. \benchmarkname{} turns shopping from understanding a single query into recovering requirements split across three sources, two gated behind tools; the agent begins with a partial view of the user's intent and must acquire the remainder through search, profile access, and clarification. Hiddenness enters on two independent axes: \emph{source}, when a requirement lives behind a tool (persona, clarification), and \emph{expression}, when the visible query states a need only implicitly. Figure~\ref{fig:overview} walks through one task end-to-end, from the three sources through the agent's tool use to source-tagged rubric scoring (full case in App.~\ref{sec:appendix_case}).

\begin{figure*}[t]
\centering
\definecolor{cbpurple}{HTML}{CC79A7}
\newcommand{\rqbul}[1]{{\color{#1}\footnotesize$\blacktriangleright$}~}
\tcbset{
  ovbox/.style={colframe=black!45, colback=white, colbacktitle=black!10, coltitle=black,
    fonttitle=\scriptsize\bfseries, fontupper=\scriptsize, boxrule=0.5pt, arc=1.5pt,
    titlerule=0.3pt, left=2.5pt, right=2.5pt, top=2pt, bottom=2pt},
}

\begin{tcbraster}[raster columns=3, raster equal height=rows, raster column skip=2.2mm, ovbox]
\begin{tcolorbox}[title={(a)~Task: intent hidden in three sources}]
  \textcolor{cbblue}{\textbf{Query}}~\textit{(visible)} --- implicit cues:\\
  \rqbul{cbblue}\texttt{Connector = micro-USB}\\
  \rqbul{cbblue}\texttt{Mounting = Tabletop}\\
  \rqbul{cbblue}review: ``standing up or lying down''\\[2pt]
  \textcolor{cborange}{\textbf{Persona}}~\textit{(hidden $\cdot$ \texttt{get\_user\_profile})}:\\
  \rqbul{cborange}\texttt{Color = Black}\\
  \rqbul{cborange}\texttt{Device = Smartphones}\\[2pt]
  \textcolor{cbgreen}{\textbf{Clarification}}~\textit{(hidden $\cdot$ \texttt{ask\_user})}:\\
  \rqbul{cbgreen}\texttt{avg\_rating $\geq$ 3.5}\\
  \rqbul{cbgreen}\texttt{Connectivity = USB}
\end{tcolorbox}
\begin{tcolorbox}[title={(b)~How the agent interacts}]
  \centering
  \begin{tikzpicture}[font=\scriptsize, >={Stealth[length=4.5pt,width=4pt]},
    n/.style={rounded corners=2.5pt, draw=black!55, line width=0.6pt, inner sep=2.5pt,
              minimum width=41mm, minimum height=5.5mm, align=center, font=\scriptsize}]
    \node[n, fill=cbpurple!18] (sh) {Shopper};
    \node[n, fill=cbblue!20, below=6.5mm of sh] (ag) {Agent};
    \node[n, fill=cbgreen!16, below=6.5mm of ag] (env) {Shop sandbox};
    \draw[->, draw=black!60] ([xshift=-13mm]sh.south) -- ([xshift=-13mm]ag.north);
    \draw[->, draw=black!60] ([xshift=13mm]ag.north) -- ([xshift=13mm]sh.south);
    \node[font=\tiny, fill=white, inner sep=0.6pt] at ($(sh)!0.5!(ag)$) {\texttt{ask\_user} / reply};
    \draw[->, draw=black!60] ([xshift=-13mm]ag.south) -- ([xshift=-13mm]env.north);
    \draw[->, draw=black!60] ([xshift=13mm]env.north) -- ([xshift=13mm]ag.south);
    \node[font=\tiny, fill=white, inner sep=0.6pt] at ($(ag)!0.5!(env)$) {action / observation};
  \end{tikzpicture}\\[3pt]
  {\scriptsize The agent must \emph{actively} fetch the profile and ask the\\ shopper. API sandbox: 10 tools, $\leq$100 steps, one product.}
\end{tcolorbox}
\begin{tcolorbox}[title={(c)~Source-tagged typed-rubric scoring}]
  recommended product: \texttt{B07DJB5F29}\\[2pt]
  \textcolor{cbblue}{\textbf{Query}}~(8/8):\\
  \rqbul{cbblue}\texttt{attribute\_match} micro-USB~\ding{51}\\
  \rqbul{cbblue}\texttt{entity\_match} Wireless Charger~\ding{51}\\
  \rqbul{cbblue}\texttt{review\_opinion} standing or lying~\ding{51}\\[2pt]
  \textcolor{cborange}{\textbf{Persona}}~(2/2):\\
  \rqbul{cborange}\texttt{attribute\_match} Color=Black~\ding{51}\\[2pt]
  \textcolor{cbgreen}{\textbf{Clarification}}~(2/2):\\
  \rqbul{cbgreen}\texttt{numeric\_range} rating$\geq$3.5~\ding{51}\\
  \rqbul{cbgreen}\texttt{attribute\_match} USB~\ding{51}\\[2pt]
  \textit{Correct = exact product match \textbf{or} all rubrics satisfied.}
\end{tcolorbox}
\end{tcbraster}

\vspace{1.5mm}

\begin{tcolorbox}[ovbox, title={Worked example: \CaseExampleModel{} recovers the hidden intent, then recommends one product (steps run top to bottom; 32 tool calls in all)}]
\setlength{\tabcolsep}{4pt}\renewcommand{\arraystretch}{1.1}
\begin{tabularx}{\linewidth}{@{}l >{\raggedright\arraybackslash}p{0.40\linewidth} >{\raggedright\arraybackslash}X@{}}
\textbf{Step\,$\downarrow$} & \textbf{Agent} & \textbf{Environment / User} \\
\midrule
1 & \textcolor{cbblue}{\texttt{get\_user\_profile}} & \textcolor{cbgreen}{returns \texttt{Color=Black}, \texttt{Device=Smartphones}} \\
2--6 & \textcolor{cbblue}{\texttt{ask\_user} $\times 5$ (5th triggers the slot)} & \textcolor{cbpurple}{User: ``at least 3.5 stars''} \emph{(unlocks \texttt{avg\_rating$\geq$3.5})} \\
7--11 & \textcolor{cbblue}{\texttt{search\_products}, \texttt{get\_product\_details}, \texttt{get\_product\_review\_stats}} & \textcolor{cbgreen}{\texttt{B07DJB5F29}: micro-USB, Tabletop, Black, USB; rating 3.7} \\
12--29 & \textcolor{cbblue}{\texttt{get\_review\_content} (mine reviews; rule out alternatives)} & \textcolor{cbgreen}{review: ``charges standing up or lying down; angle adjustable''} \\
30--32 & \textcolor{cbblue}{\texttt{search\_products}, then \texttt{recommend\_product}(\texttt{B07DJB5F29})} & \textcolor{black!75}{every rubric satisfied $\Rightarrow$ correct} \\
\end{tabularx}
\end{tcolorbox}

\caption{\benchmarkname{} on one task (wireless charger). \textbf{Top:} (a)~the full requirement is split across three sources: a visible query with implicit cues, a hidden persona behind \texttt{get\_user\_profile}, and hidden clarification behind \texttt{ask\_user}; (b)~the agent interacts with the shopper and a sandboxed shop to recover them; (c)~the recommended product is graded against typed rubrics tagged by source (a case is correct on exact product match or full rubric satisfaction). \textbf{Bottom:} an abridged trajectory of the recovery. Source colors (query in blue, persona in orange, clarification in green) carry across panels.}
\label{fig:overview}
\end{figure*}

\paragraph{Environment.} The agent operates in an API sandbox rather than a browser. This separates the shopping-specific work (finding information and checking requirements) from web-page navigation, and enables deterministic replay with fixed tool behavior. The tool set contains \ToolCount{} tools spanning catalog search and inspection (search, filtering, product details, review evidence), user interaction (profile access, clarification), and decision support (lightweight calculation, final recommendation). Each run is capped at 100 tool steps (one call per step) and 10 clarification turns, and the agent must terminate by recommending exactly one product. Recovering the scattered intent requires sustained multi-step acquisition, making each episode a long-horizon trajectory.

\paragraph{Rubrics and metrics.} Evaluation uses typed rubrics rather than a single label. Each carries a type (e.g., \texttt{attribute\_match}, \texttt{numeric\_range}, \texttt{review\_opinion}) and a source tag (query, persona, clarification), with the six types detailed in Section~\ref{sec:construction}. Our primary metric is accuracy: exact product match or full rubric satisfaction. We additionally report rubric satisfaction by source, finished rate, and tool-use statistics. This casts shopping evaluation as a joint test of retrieval, profile use, clarification, review verification, and selection.

\section{Benchmark Construction}
\label{sec:construction}

To populate this environment, we build \benchmarkname{} from a real catalog of 3.7M products and 21.4M reviews derived from Amazon Reviews 2023~\citep{amazonreviews2024}, concentrated in electronics-adjacent categories. The valid split contains \ValidSamples{} samples spanning \IntentCount{} intents, from product search and rating quality to coupon budgeting and review-driven shopping; of \TotalSamples{} generated candidates, \GenerationPassRate{}\% survive final validation (intent definitions and per-intent statistics in App.~\ref{sec:appendix_diag}).

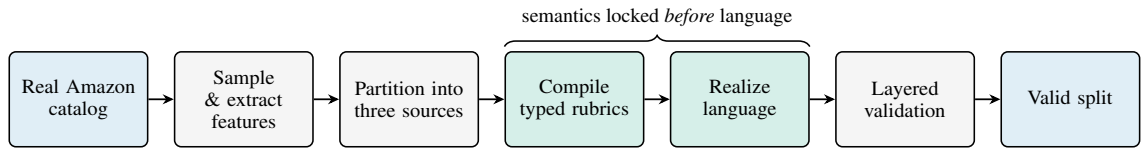
\begin{figure*}[t]
\centering
\resizebox{0.94\textwidth}{!}{
\begin{tikzpicture}[
  font=\small,
  >={Stealth[length=5pt,width=4.5pt]},
  stage/.style={rounded corners=3pt, draw, line width=0.7pt, align=center,
                fill=black!4, inner sep=4pt, text width=19mm, minimum height=15mm},
  node distance=4mm,
]
\node[stage, fill=cbblue!12] (cat)  {Real Amazon catalog};
\node[stage, right=of cat]  (feat) {Sample \& extract features};
\node[stage, right=of feat] (part) {Partition into three sources};
\node[stage, right=of part, fill=cbgreen!16] (rub)  {Compile typed rubrics};
\node[stage, right=of rub,  fill=cbgreen!16] (real) {Realize language};
\node[stage, right=of real] (val)  {Layered validation};
\node[stage, right=of val, fill=cbblue!12] (split) {Valid split};
\foreach \a/\b in {cat/feat,feat/part,part/rub,rub/real,real/val,val/split}{%
  \draw[->,line width=0.8pt] (\a) -- (\b);}
\draw[decorate, decoration={brace, amplitude=5pt, raise=2pt}, line width=0.7pt]
  (rub.north west) -- (real.north east)
  node[midway, above=7pt, font=\footnotesize] {semantics locked \emph{before} language};
\end{tikzpicture}}
\caption{Construction pipeline. \benchmarkname{} starts from a real product catalog, splits requirements across query, persona, and clarification, fixes the typed, source-tagged rubrics before any text is written, and validates every sample before it enters the valid split.}
\label{fig:construction_pipeline}
\end{figure*}

\paragraph{Controlled pipeline (RQ3).} Each sample is built by a seven-stage pipeline (Figure~\ref{fig:construction_pipeline}; details in App.~\ref{sec:appendix_validation}). The central design principle is that \emph{the correct answer is fixed before any text is generated}. Once a rubric's id, type, field, expected value, and source are set in code, later LLM calls only verbalize them and cannot alter the correctness criteria: the LLM controls surface realization, not the underlying requirements. Letting an LLM author the evaluation criteria would build its noise into the ground truth; locking them in code removes that variance at the source.

\paragraph{Three-source split (RQ2).} Extracted features are split into query-visible, persona-hidden, and clarification-hidden requirements. The valid split carries \RubricCount{} rubrics, averaging \AvgRubrics{} per sample: \QueryRubricCount{} from the query and \PersonaRubricCount{} each from persona and clarification. Each hidden requirement adds a sub-goal reachable only through tools, lengthening the horizon the agent must cover. Persona constraints are stated preferences kept in a dedicated profile field, separated from the unscored profile context to prevent leakage, and intended to be consulted rather than inferred.

\paragraph{Typed, source-tagged rubrics.} Each rubric carries a type and a source tag (Table~\ref{tab:rubric_types}). The type shows which kind of requirement failed (identification, attribute, numeric, exclusion, budget, or review judgment); the source shows whether the requirement was visible or hidden. Because each expected value is fixed in code, the judge verifies a specific target rather than grading freely, and therefore cannot drift.

\begin{table}[t]
\centering
\small
\setlength{\tabcolsep}{4pt}
\renewcommand{\arraystretch}{1.25}
\begin{tabularx}{\columnwidth}{@{}l Y@{}}
\toprule
\textbf{Type} & \textbf{Checks (example)} \\
\midrule
\texttt{attribute\_match}   & exact attribute value (\texttt{Color}=Black) \\
\texttt{numeric\_range}     & numeric bound (\texttt{average\_rating}\,$\geq$\,3.5) \\
\texttt{entity\_match}      & identity phrase in title (``Wireless Charger'') \\
\texttt{negative\_attribute}& an excluded value (e.g., not waterproof) \\
\texttt{budget\_match}      & post-voucher total within budget \\
\texttt{review\_opinion}    & opinion grounded in reviews (``standing up or lying down'') \\
\bottomrule
\end{tabularx}
\caption{The six rubric types in \benchmarkname{}. Each rubric also carries a source tag (query / persona / clarification).}
\label{tab:rubric_types}
\end{table}

\paragraph{Validation by design.} Validation is part of how the benchmark is defined, not a post-hoc filtering step. We lock each rubric before any text is generated. Automated checks keep each requirement on its assigned source and prevent leakage: a hidden value may not appear verbatim in the query. A generation check confirms that the query, persona, and clarification together support their rubrics and that the target product satisfies all of them. Clarification is deterministic: each slot has a list of trigger keywords and a fixed answer, so a missed clarification is attributable to the agent rather than to user-side stochasticity. App.~\ref{sec:appendix_validation} gives validation examples and the exact-match cross-check.

\section{Experiments}
\label{sec:results}

\paragraph{Protocol.} All results are measured on the \ValidSamples{} valid samples. The prediction phase uses the tool environment of Section~\ref{sec:task_env} (at most 100 tool steps and 10 clarification turns); the evaluation phase uses \JudgeModel{} as the unified judge. A case is correct if the agent recommends the exact target product or satisfies every rubric. This dual-path definition yields a user-facing notion of correctness and enables cross-checking the rubric score.

\paragraph{Reliability signals.} Three design choices keep grading reliable: each rubric has a fixed expected value, clarification is deterministic rather than LLM-simulated, and correctness is tracked along both the exact-match and the rubric path. We then audit the judge in both directions (App.~\ref{sec:judge_stability}). The judge rejects a correct, exact-target recommendation at most once per model. In the other direction, for one model, we re-examined every case in which the judge accepted a non-target product as correct (161 rubrics) and identified no genuine errors. The benchmark is therefore hard because of the task, not because the judge is noisy.

\paragraph{Overall performance.} Table~\ref{tab:main_results} reports headline results: overall accuracy spans \WorstAccuracy{}--\BestAccuracy{}\% under a shared environment and judge. This wide spread indicates that \benchmarkname{} separates models rather than saturating near the ceiling. \texttt{claude-opus-4-6} leads on accuracy, \texttt{gpt-5.4} on rubric satisfaction, and \texttt{kimi-k2.6} nearly matches \texttt{gpt-5.4} with a deeper, ask-late style; the differences are behavioral as well as numerical.

\begin{table*}[t]
\centering
\small
\begin{threeparttable}
\setlength{\tabcolsep}{6pt}
\begin{tabular}{l
                S[table-format=2.1]
                S[table-format=2.1]
                S[table-format=3.1]
                S[table-format=2.1]
                S[table-format=2.1]}
\toprule
Model & {Acc.} & {Rubric Sat.} & {Finish} & {Avg Calls} & {Median} \\
\midrule
Opus-4.6 & 57.1 & 76.6 & 84.6 & 33.9 & 16.0 \\
GPT-5.4 & 47.0 & 81.0 & 99.4 & 25.1 & 20.0 \\
Kimi K2.6 & 46.4 & 73.4 & 88.7 & 38.2 & 28.0 \\
GPT-5 & 39.6 & 76.1 & 97.7 & 24.4 & 15.0 \\
MiniMax M2.7 & 36.6 & 73.5 & 95.6 & 31.3 & 21.0 \\
GPT-5 mini & 31.6 & 72.7 & 100.0 & 15.5 & 12.5 \\
Qwen3-30B-A3B & 19.5 & 59.0 & 100.0 & 18.3 & 16.0 \\
\bottomrule
\end{tabular}
\begin{tablenotes}[flushleft]\footnotesize
\item Short names follow Table~\ref{tab:inference_config}. Accuracy is defined as exact product match or full rubric satisfaction.
\end{tablenotes}
\caption{Main results on the \ValidSamples{} valid \benchmarkname{} samples.}
\label{tab:main_results}
\end{threeparttable}
\end{table*}

\paragraph{Hidden intent is harder than visible intent (RQ2).} Table~\ref{tab:by_source} breaks rubric satisfaction down by source. For every model, query-visible constraints are easier than persona and clarification: \texttt{gpt-5.4} reaches 88.1\% on query but only 69.8\% and 70.9\% on persona and clarification. Almost every model reads the profile, so the deficit lies in using hidden information rather than retrieving it: a skill query-visible benchmarks cannot measure.

\begin{table}[t]
\centering
\small
\setlength{\tabcolsep}{4pt}
\begin{tabular}{lccc}
\toprule
Model & Query & Persona & Clar. \\
\midrule
Opus-4.6 & 80.6 & 72.8 & 68.6 \\
GPT-5.4 & 88.1 & 69.8 & 70.9 \\
Kimi K2.6 & 79.9 & 63.2 & 64.1 \\
GPT-5 & 85.7 & 58.6 & 64.9 \\
MiniMax M2.7 & 83.5 & 57.3 & 59.9 \\
GPT-5 mini & 82.1 & 56.6 & 60.2 \\
Qwen3-30B-A3B & 68.8 & 43.1 & 45.4 \\
\bottomrule
\end{tabular}
\caption{By-source rubric satisfaction (\%). Query constraints are consistently easier than persona and clarification constraints.}
\label{tab:by_source}
\end{table}

\paragraph{Implicit requirements drive difficulty.} The number of requirements stated only implicitly in the query predicts difficulty better than the raw number of requirements (Figure~\ref{fig:difficulty}; full buckets in App.~\ref{sec:appendix_diag}). Going from 0--1 such requirements to 3+ nearly halves the seven-model average accuracy ($51.8\%\!\to\!27.2\%$), whereas adding explicitly stated requirements does not.


\paragraph{Two interaction styles emerge, with complementary failure modes.} Trajectories reveal two strategies (App.~\ref{sec:appendix_diag}, Table~\ref{tab:behavior}). \emph{Deep-search, ask-late} models (Opus, Kimi, MiniMax) run 15--17 searches and place their first \texttt{ask\_user} past the halfway point, asking the user in only 10--20\% of tasks; \emph{shallow-search, ask-early} models (the GPT family) ask almost immediately (\texttt{gpt-5.4} in every task, at position 0.06). The two styles trade off: deep searchers verify more but exhaust the 100-step budget more often (Opus finishes only 84.6\%), while early-askers almost always finish yet fare no better on hidden requirements.

\paragraph{Failures concentrate on review evidence (RQ1).} Pooling the 12{,}467 failed rubrics localizes where agents fail. By volume, 81.5\% are \texttt{attribute\_match} (the most common type), but by rate \texttt{review\_opinion} has the lowest satisfaction of any type (33.7--67.4\%): extracting and aggregating evidence from unstructured reviews, rather than matching structured attributes, is the hardest sub-skill. Per-intent accuracy (App.~\ref{sec:appendix_diag}) corroborates this picture: product search, rating quality, and coupon budgeting are easiest; knowledge-reasoning, review-driven, and feature-combination tasks are among the hardest.

\section{Further Analysis}
\label{sec:robustness}

\paragraph{Validation filtering.} Of 921 generated dataset candidates, 662 pass validation; App.~\ref{sec:camera_ready_appendix} reports the five rejection categories and their counts. Reweighting final per-intent accuracies to the intent distribution of all candidates changes accuracy by at most 0.5 points and preserves the model ranking. Because rejected candidates are invalid and were not evaluated, their counterfactual difficulty cannot be measured.

\paragraph{Clarification sensitivity.} Each of the 1,326 clarification slots has 7--10 trigger terms (7.17 on average), and all slots are mechanically reachable. In a model-stratified audit of 651 semantically well-targeted questions asked while a requirement remained hidden, 24 (3.7\%) fail to activate the intended slot.

\paragraph{Review-evidence failures.} Among 602 failed \texttt{review\_opinion} rubrics, 226 occur in runs without a valid recommendation. Of the remaining 376, 186 retrieve some review text for the recommended product, 164 search but obtain no matching text, and 26 do not search its reviews; none fail because of an invalid review-tool call. Of 104 failures with retrieved review text that we audited, only 28 contained supporting evidence. Evidence acquisition is thus the dominant observed bottleneck, while evidence integration remains a smaller failure category.

\paragraph{Call-budget sensitivity.} On the recorded trajectories, only 225 of 4,634 runs (4.9\%) reach the 100-agent-iteration loop cap. Executed calls 76--100 add 0.57 accuracy points on average and at most 1.5 points for any model; truncating trajectories after 50 calls lowers accuracy by 3.6 points for Opus and 4.0 for Kimi, without changing the ranking. Thus, the 100-call limit rarely binds in aggregate, whereas 50 calls would disproportionately truncate deep-search behavior.

\paragraph{Exact versus judged correctness.} Of 1,838 correct predictions, 1,755 (95.5\%) exactly recover the target and are correct independently of the judge; 83 (4.5\%) are non-exact recommendations judged to satisfy every rubric. The exact path also rescues four target predictions with at least one judge error. For \texttt{gpt-5.4}, manual verification of all 17 non-exact correct predictions and their 161 rubrics found no false positives. Judge dependence is therefore limited to a small subset of correct predictions, although the audit does not establish perfect reliability for every model.

\section{Discussion}

\paragraph{Shopping as a stress test for long-horizon agency.}\looseness=-1 Shopping assistants highlight a central weakness of current agent models: the difficulty is not only finding relevant products, but sustaining a coherent decision process over many steps. \benchmarkname{} turns this everyday scenario into a long-horizon benchmark by allowing up to 100 tool calls for intent acquisition, evidence verification, and final commitment. The resulting failures are revealing: models often retrieve useful profile information without incorporating it into their decisions; exhaustive search strategies consume the step budget without converging; early-commitment strategies choose products before verification; and review evidence remains difficult because it must be accumulated and reasoned over across the trajectory. The sharp performance drop from $51.8\%\!\to\!27.2\%$ as implicit requirements accumulate suggests that current agents are better at local tool use than at global trajectory-level reasoning. Within the evaluated shopping setting, these properties make the benchmark a practical stress test for long-horizon agency~\citep{deepplanning2026}.

\paragraph{From product matching to requirement reasoning.} \benchmarkname{} is challenging not because it artificially enlarges the search space, but because it captures a core difficulty of real shopping agents: user requirements are latent, distributed, and evidence-dependent. Some requirements are stated explicitly in the query, others must be retrieved from the profile or elicited through dialogue, many are expressed as implicit clues rather than literal product attributes, and some can only be verified through review evidence. Solving the task therefore requires an agent to recover hidden intent, maintain it over a long tool-use trajectory, and connect each requirement to the right form of evidence before committing. Our by-source, difficulty, and failure analyses in Section~\ref{sec:results} indicate that the observed performance gaps track these intended sources of complexity while grader noise is minimal: no model has more than one correct answer falsely rejected by the judge.

\paragraph{Systematic error patterns.} The failure modes above are systematic rather than anecdotal. Per-requirement failure rates rise monotonically with how deeply a requirement is hidden, roughly doubling from query (12--31\%) to persona (27--57\%) and clarification (29--55\%). Failures concentrate on evidence-bearing checks, with \texttt{review\_opinion} rubrics satisfied least often among the six types. The patterns persist across interaction styles: deep searchers more often exhaust the step budget, while early-committing models almost always finish yet are no better at satisfying hidden requirements. The remaining headroom thus lies in trajectory-level capabilities (proactive information acquisition, sustained constraint integration, calibrated stopping) rather than in stronger single-step execution.

\section{Conclusion}

\looseness=-1 We recast e-commerce shopping as a long-horizon agent task and present \benchmarkname{} to evaluate it: within 100 tool calls, an agent must recover requirements distributed across a query, a tool-gated profile, and scripted clarification, verify them against catalog and review evidence, and commit to a single product. Hidden, distributed intent constructs the long horizon; typed, source-tagged rubrics tie each failure to a requirement and its source; and an automated yet validated pipeline with an audited judge keeps the evaluation reproducible. Across seven models, overall accuracy ranges from \WorstAccuracy{}\% to \BestAccuracy{}\%, and rubric satisfaction degrades from visible to hidden requirements: integrating scattered intent over a long trajectory remains an open challenge. We believe \benchmarkname{} provides a demanding testbed and concrete diagnostics for building deployable shopping agents. We release the benchmark data together with the full dataset-construction, prediction, and evaluation harness at \url{https://github.com/Morizeyao/EComAgentBench_}.\label{body:end}

\section*{Limitations}
\benchmarkname{} is deliberately scoped as a reproducible diagnostic benchmark, which implies several limitations. First, the released split draws on a bounded slice of Amazon Reviews 2023~\citep{amazonreviews2024}: products concentrate in electronics-adjacent categories, and all tasks are in English on a single marketplace. The reported accuracies therefore do not estimate performance across e-commerce domains. Extending the framework to another domain requires adapting and validating its attribute mapping. Second, every task ends in one recommended product, leaving multi-item carts, comparison dialogues, and post-purchase support to future work. Third, the user side is fully scripted for reproducibility: personas contain synthetic stated preferences (possible generation biases are discussed in the Ethics Statement), and a deterministic matcher reveals a hidden requirement only when the agent asks a matching clarification question. This is a deliberate trade-off: the same question always receives the same response, which makes failures attributable to the evaluated agent. An LLM judge or semantic matcher would add model/version dependence, calibration choices, and another source of evaluation uncertainty; the deterministic matcher nevertheless does not capture every natural paraphrase.

Frozen trajectories retain tool-call counts but not provider-reported token usage, caching, retries, or latency; future harness versions will log these quantities for compute-normalized cost and latency comparisons.
Finally, non-exact recommendations rely on an LLM judge, so residual judge error cannot be fully excluded.

\section*{Ethics Statement}
\paragraph{Data and licensing.} The product catalog is constructed from Amazon Reviews 2023~\citep{amazonreviews2024}, a publicly released research dataset. We use product metadata and review text strictly for research evaluation, consistent with the dataset's research-use terms. We release only the derived benchmark and product index necessary for reproduction, under permission from the dataset authors.

\paragraph{No personally identifiable information.} The user personas in \benchmarkname{} are synthetically generated and do not correspond to real individuals. The benchmark does not attempt to re-identify any real user, and persona fields are produced to support hidden-intent evaluation rather than to model any specific person. This is a deliberate ethical advantage over evaluations built on real user behavioral logs.

\paragraph{Bias and representativeness.} Because personas are model-generated, they may inherit biases of the generating model, such as skewed demographic, occupational, or lifestyle distributions. In addition, the released catalog is concentrated in electronics-adjacent categories and therefore does not represent all shoppers or product domains. Results should be read as diagnostic signals about hidden-intent acquisition rather than as estimates of real-world shopping success across the full population. These boundaries are stated in the Limitations.

\paragraph{Evaluated models.} We evaluate publicly available commercial and open-weight language models through their respective APIs, in accordance with their terms of use, and report only aggregate evaluation metrics.

\bibliography{refs}

\clearpage
\appendix
\section{Model and Inference Configuration}
\label{sec:appendix_inference}

We evaluate seven representative models: \texttt{gpt-5.4}, \texttt{gpt-5}, \texttt{gpt-5-mini}, \texttt{claude-opus-4-6}, \texttt{moonshotai/kimi-k2.6}, \texttt{minimax/minimax-m2.7}, and \texttt{qwen3-30b-a3b-instruct-2507}. Table~\ref{tab:inference_config} summarizes the model-specific inference settings used in evaluation, with shared defaults applied across runs.

\begin{table*}[t]
\centering
\small
\begin{threeparttable}
\setlength{\tabcolsep}{6pt}
\begin{tabularx}{\textwidth}{l Y Y}
\toprule
Model & Provider & Reasoning \\
\midrule
GPT-5.4 & OpenAI & medium reasoning \\
GPT-5 & OpenAI & medium reasoning \\
GPT-5 mini & OpenAI & medium reasoning \\
Opus-4.6 & Anthropic & provider default \\
Kimi K2.6 & OpenRouter & thinking budget 2048 \\
MiniMax M2.7 & OpenRouter & thinking budget 2048 \\
Qwen3-30B-A3B & OpenRouter & not available \\
\bottomrule
\end{tabularx}
\begin{tablenotes}[flushleft]\footnotesize
\item Shared defaults: \texttt{temperature=1.0} and \texttt{max\_tool\_steps=100} (at most 10 clarification turns).
\item Model mapping: Qwen3-30B-A3B = \texttt{qwen3-30b-a3b-instruct-2507} (the instruction-tuned MoE with 30B total / 3B active parameters).
\end{tablenotes}
\caption{Per-model inference settings used during the prediction phase.}
\label{tab:inference_config}
\end{threeparttable}
\end{table*}

\section{Additional Diagnostics}
\label{sec:appendix_diag}

This appendix reports the core benchmark statistics (Table~\ref{tab:benchmark_stats}) and complements Section~\ref{sec:results} with the full difficulty-bucket breakdown, the interaction-behavior table and scatter, and the per-intent accuracy matrix.

\begin{table}[t]
\centering
\footnotesize
\begin{tabularx}{\linewidth}{@{}p{0.40\linewidth}X@{}}
\toprule
Statistic & Value \\
\midrule
Catalog products & 3{,}721{,}595 \\
Catalog reviews & 21{,}351{,}983 \\
Total generated samples & \TotalSamples{} \\
Valid benchmark samples & \ValidSamples{} \\
Generation pass rate & \GenerationPassRate{}\% \\
Intents & \IntentCount{} \\
Tools & \ToolCount{} \\
Rubrics & \RubricCount{} \\
Rubric types & \RubricTypeCount{} \\
Average rubrics / valid sample & \AvgRubrics{} \\
Rubric source split & query \QueryRubricRatio{}\%, persona \PersonaRubricRatio{}\%, clarification \ClarificationRubricRatio{}\% \\
\bottomrule
\end{tabularx}
\caption{Core statistics of the current \benchmarkname{} benchmark, computed from the current benchmark release.}
\label{tab:benchmark_stats}
\end{table}

\begin{table}[t]
\centering
\footnotesize
\setlength{\tabcolsep}{3pt}
\begin{tabular}{lccc}
\toprule
Bucket & Count & 7-model Avg. Acc. & gpt-5.4 \\
\midrule
Rubrics 8--9 & 288 & 34.8 & 43.1 \\
Rubrics 10 & 174 & 43.5 & 49.4 \\
Rubrics 11+ & 200 & 43.4 & 50.5 \\
Implicit 0--1 & 215 & 51.8 & 59.1 \\
Implicit 2 & 335 & 36.1 & 44.5 \\
Implicit 3+ & 112 & 27.2 & 31.2 \\
Has review\_opinion & 95 & 36.1 & 44.2 \\
Has budget\_match & 33 & 45.9 & 60.6 \\
\bottomrule
\end{tabular}
\caption{Difficulty buckets from the \benchmarkname{} summaries. The number of implicitly stated query requirements is a much stronger difficulty driver than the raw rubric count.}
\label{tab:difficulty_buckets}
\end{table}

\begin{figure}[t]
\centering
\includegraphics[width=\columnwidth]{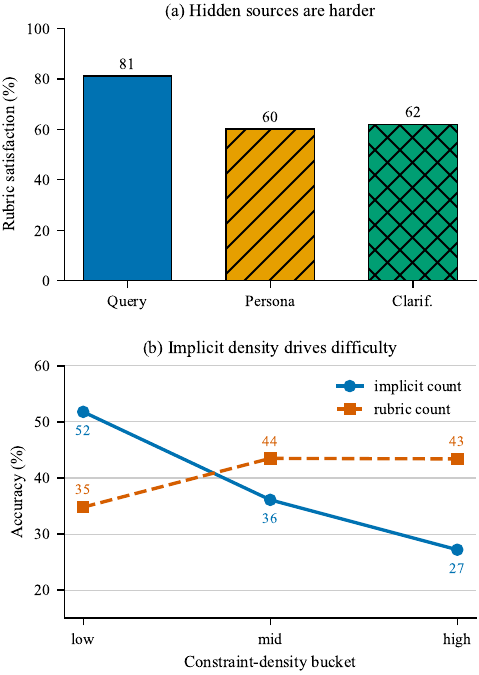}
\caption{(a)~Hidden sources (persona, clarification) stay harder than query-visible constraints. (b)~The number of implicitly stated query requirements is a stronger difficulty driver than raw rubric count.}
\label{fig:difficulty}
\end{figure}

\begin{table}[t]
\centering
\small
\setlength{\tabcolsep}{2pt}
\begin{tabular}{l
                S[table-format=2.1]
                S[table-format=2.1]
                S[table-format=1.1]
                S[table-format=1.1]
                S[table-format=1.2]
                S[table-format=3.1]}
\toprule
Model & {Search} & {Detail} & {Rev.} & {Ask} & {1st-ask} & {Finish} \\
\midrule
Opus-4.6     & 16.8 & 7.7  & 4.0 & 0.3 & 0.50 & 84.6 \\
GPT-5.4      & 7.1  & 6.6  & 4.2 & 3.8 & 0.06 & 99.4 \\
Kimi K2.6    & 14.6 & 10.4 & 7.1 & 0.5 & 0.51 & 88.7 \\
GPT-5        & 7.6  & 5.9  & 4.6 & 2.6 & 0.18 & 97.7 \\
MiniMax M2.7 & 15.3 & 7.5  & 2.1 & 1.4 & 0.50 & 95.6 \\
GPT-5 mini   & 4.6  & 3.2  & 1.7 & 3.1 & 0.27 & 100.0 \\
Qwen3-30B-A3B & 6.8  & 3.6  & 1.3 & 2.0 & 0.42 & 100.0 \\
\bottomrule
\end{tabular}
\caption{Interaction behavior mined from trajectories: mean calls to key tools, the relative position of the first \texttt{ask\_user} (0=start, 1=end), and the finish rate. Two styles emerge---deep-search/late-ask (Opus, Kimi, MiniMax) versus shallow-search/early-ask (GPT family).}
\label{tab:behavior}
\end{table}

\begin{figure}[t]
\centering
\includegraphics[width=0.96\columnwidth]{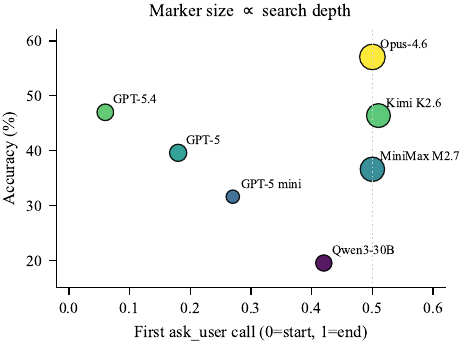}
\caption{Interaction style versus accuracy. Early-ask models (GPT family) cluster on the left; late-ask, deep-search models (Opus, Kimi, MiniMax) on the right. Marker size encodes search depth. No single style dominates---the best accuracy (Opus) pairs deep search with late clarification.}
\label{fig:behavior}
\end{figure}

Each intent category stresses a different way of expressing requirements: \emph{product search} states attributes explicitly; \emph{knowledge reasoning} paraphrases them as world-knowledge clues; \emph{use-case scenario} implies them through a concrete situation; \emph{feature combination} weaves several constraints into one request; \emph{negative constraint} excludes specific attribute values; \emph{rating quality} sets thresholds on ratings or review counts; \emph{coupon budget} constrains the post-voucher total; and \emph{review-driven} asks for experiences that must be verified in review text.

\begin{table*}[t]
\centering
\small
\setlength{\tabcolsep}{5pt}
\begin{tabular}{l *{8}{S[table-format=2.1]}}
\toprule
Model & {Prod.} & {Know.} & {Rating} & {Feat.} & {Neg.} & {Use-case} & {Coupon} & {Review} \\
\midrule
Opus-4.6     & 65.1 & 46.2 & 67.5 & 60.6 & 47.4 & 54.8 & 63.6 & 55.8 \\
GPT-5.4      & 51.8 & 43.4 & 60.0 & 39.4 & 36.8 & 48.1 & 60.6 & 44.2 \\
Kimi K2.6    & 51.8 & 38.7 & 56.2 & 43.3 & 42.1 & 45.2 & 57.6 & 45.3 \\
GPT-5        & 43.4 & 34.9 & 56.2 & 33.7 & 35.1 & 40.4 & 42.4 & 34.7 \\
MiniMax M2.7 & 49.4 & 34.0 & 42.5 & 30.8 & 40.4 & 31.7 & 42.4 & 30.5 \\
GPT-5 mini   & 42.2 & 28.3 & 40.0 & 27.9 & 24.6 & 30.8 & 33.3 & 27.4 \\
Qwen3-30B-A3B & 25.3 & 17.9 & 20.0 & 19.2 & 21.1 & 19.2 & 21.2 & 14.7 \\
\bottomrule
\end{tabular}
\caption{Per-intent accuracy (\%) for all seven models on the \ValidSamples{} valid samples. Sample counts: product search 83, knowledge reasoning 106, rating quality 80, feature combination 104, negative constraint 57, use-case scenario 104, coupon budget 33, review-driven 95.}
\label{tab:by_intent}
\end{table*}

The ordering is broadly consistent across models: product search, rating quality, and coupon budgeting are easiest on average, while knowledge reasoning, review-driven shopping, and feature combination are among the hardest---mirroring the review-evidence and multi-constraint difficulties identified in the main text.

\section{Full Benchmark Example}
\label{sec:appendix_case}

This appendix presents a representative benchmark case from the \texttt{review\_driven} intent, together with a correct \texttt{\CaseExampleModel{}} trajectory. We choose this case because it uses multiple information sources and multiple tools in a non-trivial way: the successful agent calls the profile tool, performs several clarification attempts, retrieves candidate products, checks product details, verifies review statistics, mines review evidence, and only then terminates. Catalog attribute values, including noisy ones such as the \texttt{Operating System} value ``2-COIN'' below, are preserved verbatim in the task text so that every rubric remains mechanically checkable against the catalog.

\subsection{Sample Definition}

\textbf{User query}

\begin{quote}\small
I'm looking for a Wireless Charger with reviews mentioning that it charges phone standing up or lying down and reviews stating the angle can be adjusted for convenience. It needs to be compatible with a Google LG Nexus4 on the 2-COIN operating system, include exactly the specs desktop charger, Phone Stand, fast charging, wireless charging, foldable, use the older one-way trapezoidal cord connector, and be made to rest flat on a desk rather than mounting to a wall.
\end{quote}

\textbf{User persona}

\begin{lstlisting}[language=json]
{
  "user_id": "U_40684",
  "demographics": {
    "gender": "Male",
    "age_range": "25-34",
    "location": "Chicago, IL",
    "occupation": "Logistics Coordinator",
    "education_level": "Bachelor's Degree"
  },
  "lifestyle": {
    "hobbies": ["Cooking", "Bouldering", "Playing guitar"],
    "daily_routine": "Wakes up early to hit the climbing gym before taking the train to the office.",
    "fitness_level": "Active",
    "dietary_preference": "Omnivore",
    "commute_method": "Public Transit",
    "pets": "One cat"
  },
  "shopping_habits": {
    "payment_method": "Credit Card",
    "shopping_frequency": "Bi-weekly",
    "preferred_device": "Laptop",
    "preferred_platform": "Website",
    "return_frequency": "Rarely"
  },
  "product_requirements": {
    "device_compatibility": "Smartphones",
    "color_preference": "Black"
  }
}
\end{lstlisting}

\noindent The \texttt{product\_requirements} block holds the stated preferences used as persona-source rubrics (here \texttt{device\_compatibility} and \texttt{color\_preference}); the remaining fields are benign, unscored profile context (cf.\ the persona safety whitelist in App.~\ref{sec:appendix_validation}).

\textbf{Clarification script}

\begin{lstlisting}[language=json]
{
  "clarification_slots": [
    {
      "slot_id": "cl_1",
      "linked_rubric_ids": ["r10"],
      "hidden_info": "Requires a minimum average rating of 3.5 stars.",
      "trigger_keywords": [
        "rating", "reviews", "stars", "average rating", "score", "feedback", "rated"
      ],
      "user_response": "I want something with pretty good feedback, so it should have an average rating of at least 3.5 stars.",
      "revealed": false
    },
    {
      "slot_id": "cl_2",
      "linked_rubric_ids": ["r11"],
      "hidden_info": "Requires USB connectivity technology.",
      "trigger_keywords": [
        "connectivity", "connectivity technology", "connection", "plug", "interface", "port", "cable"
      ],
      "user_response": "It definitely needs to have USB connectivity so I can plug it into my existing adapters easily.",
      "revealed": false
    }
  ],
  "default_response": "Hmm, I'm not sure about that. Could you ask me about something more specific? Like what features or specs I'm looking for?",
  "max_clarification_turns": 10
}
\end{lstlisting}

\subsection{Raw Rubrics}

\begin{lstlisting}[language=json]
[
  {
    "id": "r1",
    "type": "review_opinion",
    "field": "review",
    "expected_value": "charges phone standing up or lying down",
    "description": "charges phone standing up or lying down",
    "query_surface": "charges phone standing up or lying down",
    "info_source": "query"
  },
  {
    "id": "r2",
    "type": "review_opinion",
    "field": "review",
    "expected_value": "angle can be adjusted for convenience",
    "description": "angle can be adjusted for convenience",
    "query_surface": "angle can be adjusted for convenience",
    "info_source": "query"
  },
  {
    "id": "r3",
    "type": "attribute_match",
    "field": "details.Compatible Phone Models",
    "expected_value": "Google LG Nexus4",
    "description": "Google LG Nexus4",
    "query_surface": "Google LG Nexus4",
    "info_source": "query"
  },
  {
    "id": "r4",
    "type": "attribute_match",
    "field": "details.Operating System",
    "expected_value": "2-COIN",
    "description": "2-COIN",
    "query_surface": "2-COIN",
    "info_source": "query"
  },
  {
    "id": "r5",
    "type": "attribute_match",
    "field": "details.Mounting Type",
    "expected_value": "Tabletop Mount",
    "description": "made to rest flat on a desk rather than mounting to a wall",
    "query_surface": "made to rest flat on a desk rather than mounting to a wall",
    "info_source": "query"
  },
  {
    "id": "r6",
    "type": "attribute_match",
    "field": "details.Connector Type",
    "expected_value": "micro usb",
    "description": "older one-way trapezoidal cord connector",
    "query_surface": "older one-way trapezoidal cord connector",
    "info_source": "query"
  },
  {
    "id": "r7",
    "type": "attribute_match",
    "field": "details.Special Feature",
    "expected_value": "desktop charger, Phone Stand, fast charging, wireless charging, foldable",
    "description": "desktop charger, Phone Stand, fast charging, wireless charging, foldable",
    "query_surface": "desktop charger, Phone Stand, fast charging, wireless charging, foldable",
    "info_source": "query"
  },
  {
    "id": "r8",
    "type": "attribute_match",
    "field": "details.Compatible Devices",
    "expected_value": "Smartphones",
    "info_source": "persona"
  },
  {
    "id": "r9",
    "type": "attribute_match",
    "field": "details.Color",
    "expected_value": "Black",
    "info_source": "persona"
  },
  {
    "id": "r10",
    "type": "numeric_range",
    "field": "average_rating",
    "expected_value": {"min": 3.5},
    "info_source": "clarification"
  },
  {
    "id": "r11",
    "type": "attribute_match",
    "field": "details.Connectivity Technology",
    "expected_value": "USB",
    "info_source": "clarification"
  },
  {
    "id": "r12",
    "type": "entity_match",
    "field": "title",
    "expected_value": "Wireless Charger",
    "description": "Wireless Charger",
    "query_surface": "Wireless Charger",
    "canonical_entity": "Wireless Charger",
    "title_span": "Wireless Charger",
    "info_source": "query"
  }
]
\end{lstlisting}

\subsection{Prediction Prompt}

The prediction run uses the following system prompt for the shopping agent.

\begin{lstlisting}[basicstyle=\ttfamily\tiny]
You are a personal shopping assistant with access to a product catalog
and the shopper's profile.

Your job is to find the single product that best matches everything this
shopper needs. Their full requirements come from three sources that you
must consult:
1. The shopping query they just sent you.
2. Their saved user profile.
3. Their answers to your follow-up questions - use ask_user to probe for
   any remaining unknowns properties.

Before every tool call, write a Thought block in your response:

Thought:
- Observation: what I learned from the previous tool result
  (skip on the first turn)
- Reasoning: how this narrows down or changes my strategy
- Plan: which tool I will call next and why

Rules:
- One tool call per step.
- Recommend exactly one product by calling recommend_product. Do not
  answer in plain text.
- You have a limited of 100 steps. If you tried hard and are very near
  step limit, try to make a best guess.
- All three information sources contain parts of requirements for the
  final product.
- If search results do not match, try to broaden your search. Consider
  that the catalog uses specific attribute terms that may differ from
  the shopper's phrasing.
- This is a strict evaluation. The recommended product must satisfy
  EVERY requirement - missing even one attribute means the case is judged
  as completely wrong. Be thorough.
\end{lstlisting}

\subsection{Abridged Correct Trajectory}

The successful \texttt{gpt-5.4} run makes 32 tool calls in total, including 5 clarification attempts. Table~\ref{tab:case_traj} presents an abridged but faithful trajectory.

\begin{table*}[t]
\centering
\small
\begin{tabularx}{\textwidth}{p{0.08\textwidth}p{0.16\textwidth}X}
\toprule
Step & Tool & Summary \\
\midrule
1 & {\footnotesize\shortstack[l]{\texttt{get\_user\_}\\\texttt{profile}}} & The agent starts by retrieving hidden profile constraints, recovering \texttt{Compatible Devices = Smartphones} and \texttt{Color = Black}. \\
2 & \texttt{ask\_user} & Asks whether the ``older trapezoidal connector'' means Micro USB; no slot is triggered. \\
3 & \texttt{ask\_user} & Asks whether the charger should use Micro USB or USB-C; no slot is triggered. \\
4 & \texttt{ask\_user} & Asks whether black is mandatory or only preferred; no slot is triggered. \\
5 & \texttt{ask\_user} & Asks whether the user wants a stand rather than a flat charging pad; no slot is triggered. \\
6 & \texttt{ask\_user} & Asks whether reviews must say the charger works standing up or lying down; this triggers the rating slot and reveals the hidden clarification requirement \texttt{average\_rating >= 3.5}. \\
7 & {\footnotesize\shortstack[l]{\texttt{search\_}\\\texttt{products}}} & Searches for ``wireless charger stand foldable fast charging'' to obtain initial candidate products. \\
8--10 & {\footnotesize\shortstack[l]{\texttt{get\_product\_}\\\texttt{details}}} & Inspects multiple candidates. The key turning point is product \texttt{B07DJB5F29}, which matches \texttt{2-COIN}, \texttt{micro usb}, \texttt{Google LG Nexus4}, \texttt{Black}, \texttt{Tabletop Mount}, and the exact special-feature string. \\
11 & {\footnotesize\shortstack[l]{\texttt{get\_product\_}\\\texttt{review\_stats}}} & Verifies that \texttt{B07DJB5F29} has average rating 3.7, satisfying the clarification threshold. \\
12--16 & {\footnotesize\shortstack[l]{\texttt{get\_review\_}\\\texttt{content}}} & Searches reviews using broad and narrow review phrases. Several searches return no matches, so the agent revises its review-query phrasing. \\
17--21 & {\footnotesize\shortstack[l]{\texttt{get\_product\_}\\\texttt{details /}\\\texttt{search\_}\\\texttt{products}}} & Briefly checks alternative candidates to avoid premature commitment. These alternatives fail critical structural constraints such as connector type or compatibility coverage. \\
22--29 & {\footnotesize\shortstack[l]{\texttt{get\_review\_}\\\texttt{content}}} & Returns to \texttt{B07DJB5F29} and mines reviews with shorter keywords such as \texttt{stand}, \texttt{pad}, \texttt{positions}, and \texttt{vertically}. The retrieved review text mentions multiple positions/orientations, an adjustable stand, and the phone sitting vertically. \\
30--31 & {\footnotesize\shortstack[l]{\texttt{search\_}\\\texttt{products}}} & Performs two last broad searches to rule out a better structural match. No stronger candidate appears. \\
32 & {\footnotesize\shortstack[l]{\texttt{recommend\_}\\\texttt{product}}} & Recommends \texttt{B07DJB5F29} with reasoning that cites the exact structured matches and the review-backed evidence for adjustable upright use. \\
\bottomrule
\end{tabularx}
\caption{Abridged correct trajectory for a representative review-driven sample under \CaseExampleModel{}.}
\label{tab:case_traj}
\end{table*}

\subsection{Why the Prediction Is Correct}

The evaluation file marks this trajectory as an exact product match with zero failed rubrics. The support is distributed across both structured details and review evidence:
\begin{itemize}
  \item \textbf{Review evidence}: the judge finds review support for both required opinions, namely charging while standing up or lying down and adjustable-angle convenience.
  \item \textbf{Query-source details}: the product explicitly matches \texttt{Google LG Nexus4}, \texttt{2-COIN}, \texttt{Tabletop Mount}, \texttt{micro usb}, and the exact special-feature string.
  \item \textbf{Persona-source details}: the product matches \texttt{Compatible Devices = Smartphones} and \texttt{Color = Black}.
  \item \textbf{Clarification-source details}: the product has average rating 3.7 and connectivity technology \texttt{USB}.
\end{itemize}

This case is useful because it demonstrates both the benchmark's difficulty and its intended solution pattern: the agent must combine hidden persona information, at least one successful clarification, structured attribute verification, and review evidence retrieval before terminating.

\section{Validation and Stability Examples}
\label{sec:appendix_validation}

This appendix complements the main text by making the benchmark's stabilization logic more concrete.

\subsection{Locking the Correct Answer}

We fix what each rubric checks before any text is generated. Each rubric already has a fixed set of fields:
\begin{itemize}
  \item \texttt{id}
  \item \texttt{type}
  \item \texttt{field}
  \item \texttt{expected\_value}
  \item \texttt{info\_source}
\end{itemize}
Later steps can phrase these requirements in natural language, but cannot change what counts as correct.

\subsection{Automated Checks}

\textbf{Preventing query leaks.} The query may contain implicit clues, but hidden persona and clarification values must not appear directly in the query. Samples that break this rule are rejected before entering the valid split.

\textbf{Persona safety whitelist.} Persona generation separates product-relevant requirements from benign noise. Noise fields are restricted to safe locations such as demographics, lifestyle, and shopping habits. This reduces accidental insertion of scored information into irrelevant fields.

\textbf{Clarification completeness.} Every clarification rubric must have a clarification slot with trigger keywords and a fixed answer. If any clarification rubric lacks a slot, or a slot lacks its keywords or answer, the sample fails validation.

\subsection{Semantic Check}

After the automated checks pass, a generation-time check confirms that:
\begin{itemize}
  \item each information source supports the rubrics assigned to it,
  \item hidden values are not accidentally exposed in the wrong source,
  \item the target product satisfies the final rubric set as a whole.
\end{itemize}

This second layer is necessary because each part being individually valid does not guarantee they work together.

\subsection{LLM-Judge Stability: Errors in Both Directions}
\label{sec:judge_stability}

The dual-path correctness definition---exact product match or full rubric satisfaction---lets us check judge error in both directions, because an exact match is correct without needing the judge.

\paragraph{False negatives (judge rejects a correct answer).} \texttt{judge\_misjudged\_count} counts cases where the agent recovered the exact target product (hence necessarily correct) yet the judge scored some rubric as unsatisfied. Across the seven models it never exceeds~1:
\begin{itemize}
  \item claude-opus-4-6: 1 \quad gpt-5.4: 0 \quad moonshotai/kimi-k2.6: 1
  \item gpt-5: 1 \quad minimax/minimax-m2.7: 1 \quad gpt-5-mini: 0 \quad qwen3-30b-a3b-instruct-2507: 0
\end{itemize}
The judge thus almost never penalizes a verifiably correct recommendation.

\paragraph{False positives (judge accepts an incorrect answer).} The only correct cases that rely on the judge rather than on exact recovery are \emph{non-exact-but-correct}: the agent recommended a product other than the target, yet the judge marked all rubrics satisfied. Such cases are few---3--6\% of each model's correct cases (4.5\% pooled, 83 of 1838)---which already upper-bounds how much judge leniency could inflate accuracy. To test whether they are genuine false positives or valid equivalent products, we independently re-verified every judge-satisfied rubric of all 17 such cases for \texttt{gpt-5.4} (161 rubrics) against the predicted product's true attributes and reviews from the product catalog, followed by manual confirmation. \textbf{None were genuine false positives}: every predicted product truly satisfies all of its rubrics and is a valid alternative to the target. In four rubrics the matched catalog field was itself noisy or mislabeled---e.g., a Canon DSLR whose \texttt{Form Factor} field reads ``Compact''---and the judge correctly relied on substantive evidence (title, material, connectivity); these are catalog-data artifacts, not judge errors.

\paragraph{Takeaway.} Both error directions are small, and on the audited model the judge also handles noisy catalog fields well: it rarely rejects correct answers ($\leq$\,1 false negative per model) and accepts no incorrect ones among the non-exact-correct cases (zero false positives). We therefore conclude the benchmark is hard because of its design, not because the judge is noisy---while not claiming perfect reliability.

\section{Additional Construction and Scope Analyses}
\label{sec:camera_ready_appendix}

\subsection{Category-Level Results}

Table~\ref{tab:category_breakdown} reports categories with at least 30 tasks, covering 630 of the 662 released tasks. Pooled accuracy is the mean over all model--task pairs in a category; the model range gives the lowest and highest per-model accuracy. Absolute difficulty varies substantially across categories, but this analysis remains within the released electronics-adjacent setting and does not establish transfer to other shopping domains.

\begin{table*}[t]
\centering
\small
\begin{tabular}{lrrrr}
\toprule
Main category & Tasks & Pooled accuracy & Model range & Rubric satisfaction \\
\midrule
Computers                   & 162 & 34.8 & 17.3--56.8 & 70.9 \\
Cell Phones \& Accessories  & 139 & 47.0 & 25.2--64.0 & 78.2 \\
All Electronics             & 110 & 29.6 & 13.6--43.6 & 71.5 \\
Tools \& Home Improvement   &  58 & 48.5 & 25.9--70.7 & 74.5 \\
Amazon Home                 &  56 & 51.8 & 25.0--69.6 & 79.0 \\
Camera \& Photo             &  55 & 22.1 &  3.6--32.7 & 59.0 \\
Office Products             &  50 & 54.3 & 32.0--70.0 & 80.6 \\
\bottomrule
\end{tabular}
\caption{Category-level results for categories with at least 30 tasks. Accuracy and rubric satisfaction are percentages.}
\label{tab:category_breakdown}
\end{table*}

\subsection{Validation Filtering}

Table~\ref{tab:filtering_statistics} decomposes the 259 candidates rejected by end-to-end validation. The first three groups are broken or ungradable, leakage cases expose information intended to remain hidden, and the final group is semantically invalid. Validation thereby enforces that every requirement is recoverable from its assigned source and that the designated target satisfies its rubrics.

\begin{table}[t]
\centering
\small
\begin{tabular}{lr}
\toprule
Primary rejection reason & Count \\
\midrule
Missing query realization & 80 \\
Catalog/rubric inconsistency & 79 \\
Invalid negative constraint & 42 \\
Answer-revealing leakage & 49 \\
Unnatural, contradictory, or other & 9 \\
\midrule
Total rejected & 259 \\
\bottomrule
\end{tabular}
\caption{Primary reasons for rejecting 259 of the 921 generated candidates.}
\label{tab:filtering_statistics}
\end{table}

Because rejected candidates are not a meaningful evaluation set, their counterfactual model difficulty cannot be measured directly. As a composition check, we reweighted each model's retained-task per-intent accuracy to the intent distribution of all 921 candidates. The resulting change is at most 0.5 accuracy points for every model, and the model ranking is unchanged. This check does not rule out within-intent selection effects.


\end{document}